\documentclass[letterpaper,10pt,conference]{ieeeconf}

\IEEEoverridecommandlockouts
\usepackage[utf8]{inputenc}

\usepackage{graphicx}
\usepackage{array}
\usepackage{booktabs}
\usepackage{multirow}
\usepackage{amssymb}
\usepackage{amsmath}
\usepackage{cite}
\usepackage{flushend}
\usepackage[hidelinks]{hyperref}

\hypersetup{
  pdftitle={Evidence-Gated Task and Motion Planning with Vision-Language Models},
  pdfauthor={Tsunehiko Tanaka, Matthew Stephenson, Alistair Macvicar, Edgar Simo-Serra},
  pdfkeywords={Task and Motion Planning, Vision-Language Models, Evidence Acquisition, Feasibility Gating}
}

\title{\LARGE \bf
Evidence-Gated Task and Motion Planning \mbox{with Vision-Language Models}
}

\author{
  Tsunehiko Tanaka$^{1}$\thanks{Corresponding author: Tsunehiko Tanaka (\href{mailto:tsunehiko@fuji.waseda.jp}{tsunehiko@fuji.waseda.jp})}, Matthew Stephenson$^{2}$, Alistair Macvicar$^{2}$, and Edgar Simo-Serra$^{1}$\\
  {\normalsize ${}^{1}$Waseda University, ${}^{2}$Flinders University}
}

\begin{document}
\maketitle
\thispagestyle{empty}
\pagestyle{empty}

\begin{abstract}
Robots executing long-horizon manipulation tasks from natural-language instructions must reason about both semantic task structure and geometric feasibility.
Hierarchical approaches combining Vision-Language Models (VLMs) with Task and Motion Planning (TAMP) offer a promising solution, but VLM-generated subgoals may rely on prior knowledge without sufficient observational support.
This can lead to the omission of task-relevant subgoals or attempts to execute instructions whose requirements cannot be met in the environment.
We propose Evidence Acquisition and Feasibility Gating (EAFG), a framework for evidence-gated task and motion planning that acquires visual evidence to specify appropriate task plans and assess whether given instructions can be fulfilled.
EAFG first acquires evidence by having a VLM generate subgoals and a robot execute action sequences planned by TAMP for these subgoals.
For feasibility gating, the acquired evidence is used to decide whether to proceed with task planning, acquire further evidence, or halt.
When proceeding, the VLM generates task subgoals conditioned on the acquired evidence, and TAMP computes low-level actions.
We evaluate EAFG in three cooking scenarios involving explicitly specified required objects, underspecified instructions, and instructions requiring an absent object.
EAFG improves task completion over a VLM-TAMP baseline in the explicit and underspecified scenarios and enables appropriate halting when a required object is absent.
\end{abstract}

\begin{figure*}[t]
    \centering
    \includegraphics[width=0.9\linewidth]{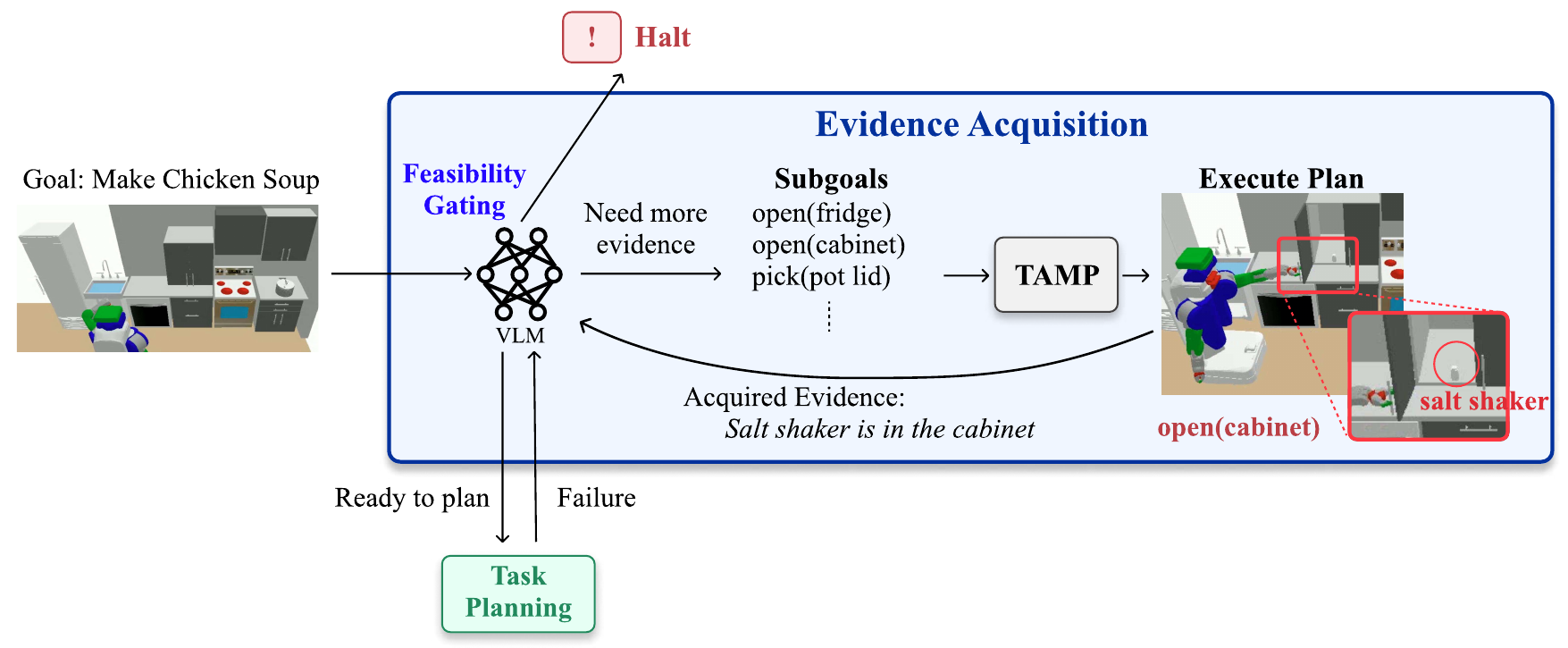}
    \caption{\textbf{Overview of Evidence Acquisition and Feasibility Gating (EAFG).} The system executes evidence-acquisition subgoals to verify goal-relevant objects and conditions, updates the evidence state from the resulting observations, and uses feasibility gating to decide whether to continue evidence acquisition, halt, or proceed to task planning.}
    \label{fig:eafg_method}
\end{figure*}

\section{Introduction}
Robots operating in everyday environments must be able to execute complex, long-horizon manipulation tasks from natural-language instructions. For example, an instruction such as ``make soup'' must be expanded into a sequence of manipulation actions involving multiple objects, including available ingredients and heating appliances. Such tasks require the robot to interpret the semantic meaning of the language instruction while satisfying geometric and kinematic constraints, such as grasp feasibility, placement stability, and collision-free reachability. Task and Motion Planning (TAMP) has therefore been widely used as a core technique for long-horizon manipulation planning, as it can generate action sequences that account for the geometric and kinematic aspects of the problem~\cite{tamp, pddlstream}. On the other hand, Vision-Language Models (VLMs) can integrate visual and linguistic information to infer commonsense procedures and semantically meaningful intermediate goals from natural-language objectives. Consequently, a hierarchical integration in which a VLM generates high-level subgoals and TAMP refines each subgoal into a geometrically feasible action sequence is a promising approach to natural-language long-horizon robot manipulation~\cite{vlm-tamp, owl-tamp}. However, the success of such systems depends on whether the generated subgoals are actually supported by observations of the environment.

A VLM may generate plausible plans based on prior knowledge without adequately checking whether observations support its assumptions about the environment. Studies have shown that VLMs can prioritize language priors over visual evidence when generating responses~\cite{lee-etal-2025-vlind, guan2024hallusionbench}. For an underspecified instruction, a VLM risks omitting task-relevant subgoals if it plans without further observation. The observations of the environment should be used to specify steps consistent with the instruction. The VLM can also generate plans for instructions whose requirements cannot be met in the environment. Such plans may lead to inefficient actions or results that fail to satisfy the instruction. The system should acquire further evidence when task-relevant requirements remain unresolved and refrain from task execution if the VLM determines that sufficient evidence for task planning cannot be acquired from the environment. Observational evidence is therefore necessary both to identify appropriate task subgoals and to assess whether the instruction can be executed.

In this paper, we propose Evidence Acquisition and Feasibility Gating (EAFG), a framework for evidence-gated task and motion planning that acquires evidence to create task plans and assess whether given instructions can be fulfilled. The acquired evidence helps the VLM specify an appropriate plan consistent with the actual environment and avoid task planning based on assumptions unsupported by observations. The VLM first generates a sequence of subgoals to acquire evidence from the environment through actions such as opening the doors of closed storage spaces. TAMP computes a sequence of actions for each subgoal and the robot executes them. The VLM then uses the acquired evidence to decide whether to proceed with task planning, acquire further evidence, or halt. If it proceeds, the VLM generates another sequence of subgoals for task execution conditioned on the acquired evidence, and TAMP computes low-level actions. We construct three cooking scenarios to evaluate EAFG: instructions with explicitly specified required objects, underspecified instructions, and instructions requiring an object absent from the environment. Our results show that EAFG outperforms a VLM-TAMP baseline in task completion rate for the explicit and underspecified scenarios. EAFG is also able to halt appropriately in the infeasible scenario, where a required object is absent from the environment.

The contributions of this work are threefold:
\begin{itemize}
    \item We introduce a framework for evidence-gated task and motion planning that acquires evidence to specify an appropriate task plan consistent with the actual environment.
    \item We propose a gating mechanism that uses acquired evidence to assess whether the given instruction can be fulfilled and decides whether to proceed with task planning, acquire further evidence, or halt execution.
    \item We design three cooking scenarios spanning instructions with explicitly specified objects, underspecified instructions, and infeasible instructions requiring objects absent from the environment. We empirically show that EAFG improves grounded task completion and infeasibility detection.
\end{itemize}

\section{Related Work}
Task and Motion Planning combines discrete symbolic reasoning with continuous motion planning to solve long-horizon manipulation tasks under geometric, kinematic, collision, grasping, and placement constraints~\cite{garrett2021integrated, pddlstream, kaelbling2011hierarchical, marc2015logic}. PDDLStream~\cite{pddlstream} extends PDDL with streams that interface with black-box procedures such as inverse-kinematics solvers, grasp and placement samplers, collision checkers, and motion planners. It coordinates finite PDDL search with stream evaluations to generate and verify the continuous parameters required by candidate manipulation plans.

Large Language Models (LLMs) and VLMs bridge open-ended language and robotic execution by providing commonsense semantics, object grounding, and spatial reasoning~\cite{pmlr-v205-ichter23a, liang2023code, 10161317, huang2023voxposer}. In planner-based systems, they translate perceptual or linguistic context into formal goals, task specifications, constraints, and symbolic representations~\cite{liu2023llmp, chen2024autotamp, guo2025castl, athalye2025pixelspredicateslearningsymbolic}. Building on these capabilities, some approaches use VLMs to guide task planning and TAMP to handle geometric and kinematic feasibility~\cite{vlm-tamp,owl-tamp}. In VLM-TAMP~\cite{vlm-tamp}, the VLM uses language and visual context to decompose a long-horizon task into semantically meaningful subgoals. These subgoals can also support acquiring evidence through actions such as opening storage spaces. However, its original formulation lacks an explicit mechanism for selecting \texttt{halt} when the instruction cannot be fulfilled.

Prior work on planning under partial observability uses information-gathering actions for object search and manipulation~\cite{li2016act, xiao2019online, kaelbling2013integrated, ss-replan, curtis2024partially}. Belief-space TAMP~\cite{kaelbling2013integrated,ss-replan,curtis2024partially} integrates information gathering and manipulation toward formally specified goals using explicit uncertainty representations. In contrast, EAFG focuses on grounding the required objects and task subgoals of underspecified natural-language instructions in acquired evidence, and on evidence-based halting when instruction requirements cannot be met. Language-guided object search uses semantic knowledge and evolving scene representations primarily to find target objects~\cite{paudel2026objectsearch, honerkamp2024language}. EAFG instead connects evidence acquisition and manipulation planning in both directions to support full-task completion. Evidence guides task-subgoal generation, while feedback from failures in manipulation planning guides further evidence acquisition.

\section{Problem Setting}

We consider long-horizon manipulation tasks specified in natural language, in which the system has access to a known manipulation domain and a TAMP backend. The system receives a natural-language goal instruction $\mathcal{G}^{\mathrm{eng}}$ and scene observations, and is required to execute an action sequence that fulfills $\mathcal{G}^{\mathrm{eng}}$. Explicitly stated requirements in $\mathcal{G}^{\mathrm{eng}}$ must be satisfied, while underspecified details should be determined from observations. Task-relevant objects could initially be hidden or absent, so the robot acquires additional evidence as needed to assess their availability. Our objective is to make the system carry out the instruction using observation-supported subgoals. It also refrains from task execution when required objects remain unverified and further inspection is judged unhelpful.

In our evaluation, transitions and observations are deterministic, $x_{t+1}=f(x_t,u_t)$ and $o_{t+1}=h(x_{t+1})$, where $x_t$ comprises symbolic facts and continuous robot and object configurations. Visible objects are assumed to be detected without error and added to the PDDL problem supplied to TAMP. Object instances that have not yet been observed and their poses are excluded from this problem. For each subgoal, TAMP searches for an action sequence $u_t$ whose continuous parameters satisfy geometric and kinematic constraints. We focus on how the VLM checks its planning assumptions against environmental observations in a setting with controlled object visibility and availability.

\begin{figure*}[t]
  \centering
  \begin{minipage}[t]{0.48\linewidth}
    \centering
    \includegraphics[width=0.8\linewidth]{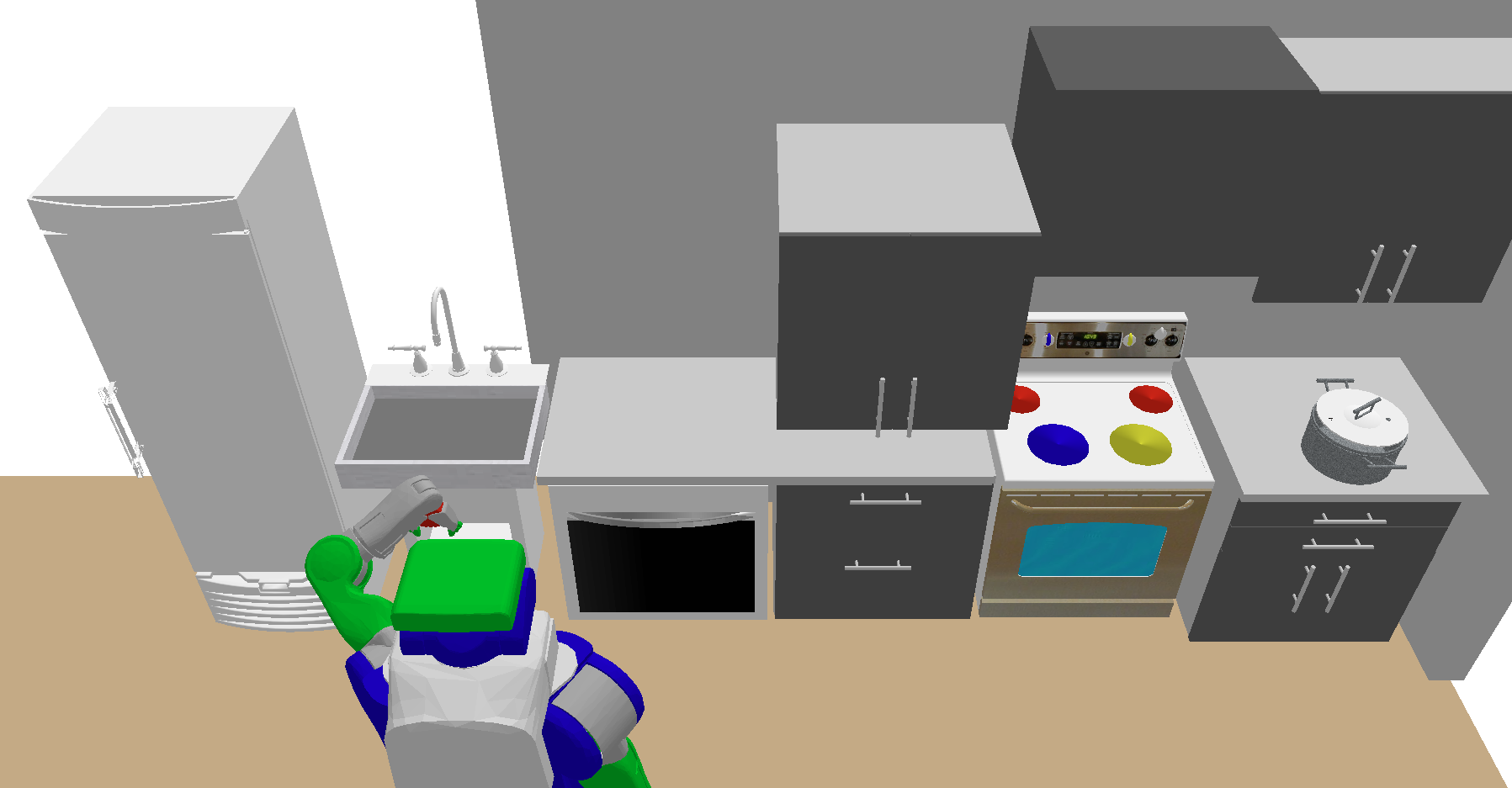}\\
    \footnotesize (a) Initial kitchen state.
  \end{minipage}
  \hfill
  \begin{minipage}[t]{0.48\linewidth}
    \centering
    \includegraphics[width=0.8\linewidth]{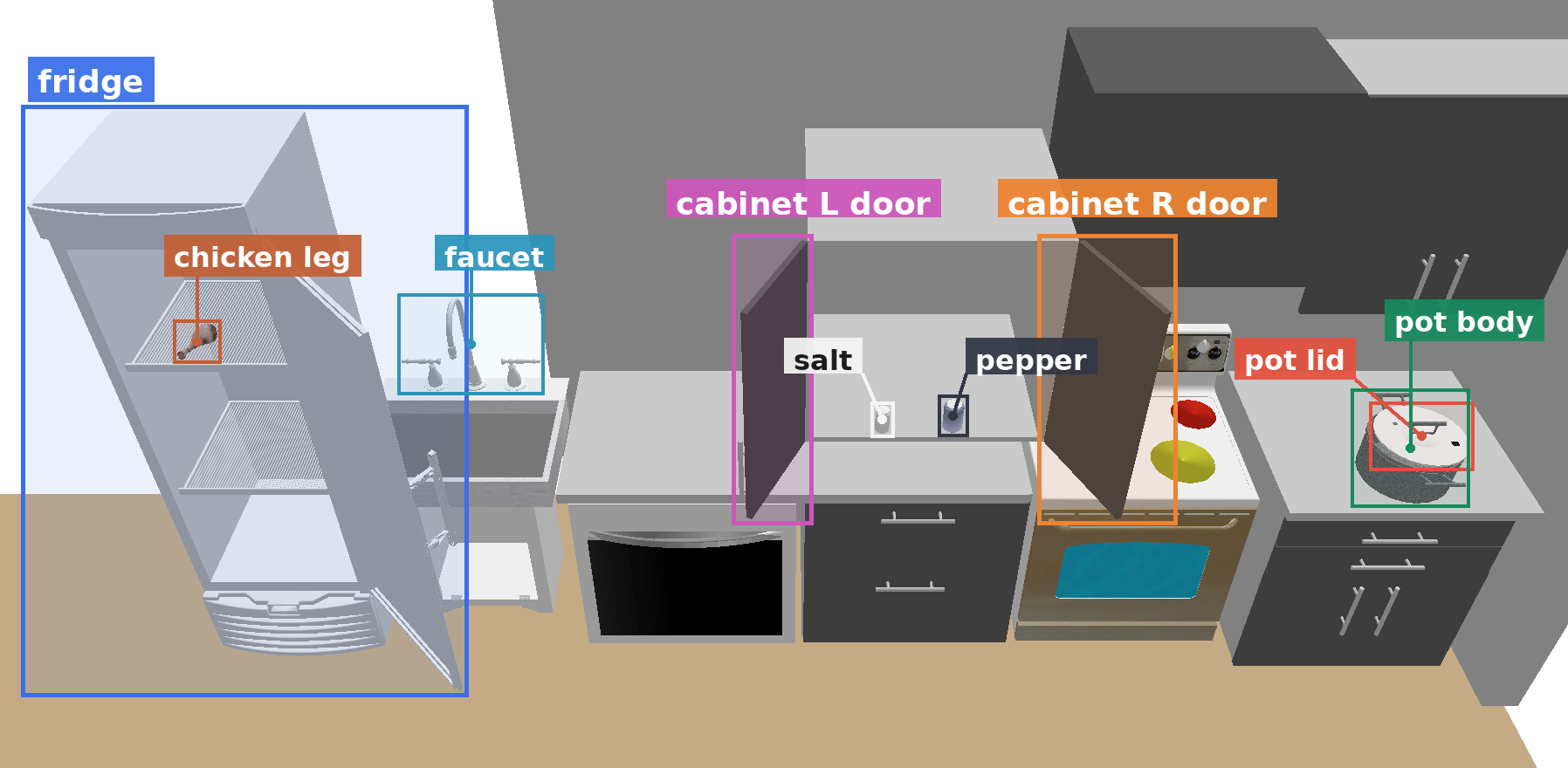}\\
    \footnotesize (b) Object layout in the kitchen.
  \end{minipage}
  \caption{Kitchen environment used in the experiments. (a) Initial state of the kitchen. (b) Layout of objects in the kitchen.}
  \label{fig:kitchen_environment}
\end{figure*}

\section{Evidence Acquisition and Feasibility Gating}
We propose Evidence Acquisition and Feasibility Gating (EAFG), a framework that acquires the observational evidence required for task planning and determines whether to proceed with task planning. EAFG uses a VLM to generate exploratory subgoals to collect evidence about goal-relevant objects and environment states. Based on the acquired evidence, the VLM decides whether to acquire further evidence, proceed with task planning, or halt execution.

\subsection{Overview}
Fig.~\ref{fig:eafg_method} summarizes the overall EAFG pipeline. EAFG takes a natural-language goal instruction $\mathcal{G}^{\mathrm{eng}}$ as input and generates a plan for acquiring evidence relevant to the goal. The VLM first proposes a sequence of evidence-acquisition subgoals
\[
    \mathcal{G}_1^{\mathrm{EA}} = \left( g_{1,1}^{\mathrm{EA}}, \ldots, g_{1,K_1}^{\mathrm{EA}} \right),
\]
where each $g_{1,k}^{\mathrm{EA}}$ denotes an exploratory subgoal, such as opening doors or cabinets or temporarily lifting occluding objects. These subgoals are used to inspect parts of the environment that could contain goal-relevant objects, rather than to directly perform the target task. Each generated evidence-acquisition subgoal is passed to a TAMP module, which attempts to instantiate it as a geometrically and kinematically feasible action sequence. The robot then executes the resulting action sequences. After each subgoal execution, the resulting observation image is stored as visual evidence.

After evidence acquisition, the VLM takes the observation images as input and outputs one of three gate statuses: proceed with task planning, acquire further evidence, or halt execution. If further evidence is needed, the VLM proposes additional evidence-acquisition subgoals, and the process repeats. The system halts if the VLM determines that the required evidence has not been obtained and that further exploration is unlikely to provide it.

Once sufficient evidence has been acquired, the system proceeds to evidence-conditioned task planning, in which the VLM uses the accumulated evidence to generate task subgoals for the original goal instruction. Each subgoal is passed to TAMP to generate an action sequence, and the robot then executes it. If TAMP cannot find a feasible action sequence, the system discards the remaining task subgoals and returns to evidence acquisition, using the failed subgoal and available failure information to guide further exploration.

\subsection{Evidence Acquisition}
\label{subsec:ea}
Evidence acquisition is designed to acquire evidence for specifying task subgoals that are consistent with the actual environment and for assessing whether the given instruction $\mathcal{G}^{\mathrm{eng}}$ can be fulfilled. It proceeds as an iterative loop, indexed by $t=1,2,\ldots$, in which the robot executes evidence-acquisition subgoals and the VLM updates its understanding of the environment from the resulting observations.

At iteration $t$, the VLM generates evidence-acquisition subgoals $\mathcal{G}_t^{\mathrm{EA}}$ from the language goal $\mathcal{G}^{\mathrm{eng}}$, current observation images, and acquired evidence to inspect potentially relevant regions. To keep the process focused on evidence acquisition, these subgoals are restricted to a manually predefined set of actions, such as opening doors and picking up objects. Irreversible or main-task actions, such as cooking, sprinkling, or heating, are excluded. This restriction keeps the generated subgoals focused on evidence acquisition until task feasibility has been assessed.

The subgoals $\mathcal{G}_t^{\mathrm{EA}}$ are passed to the TAMP module, which attempts to instantiate each subgoal as a geometrically and kinematically feasible low-level action sequence. The robot executes the subgoals in order, and after each subgoal is executed, the resulting observation image is recorded. After the execution of $\mathcal{G}_t^{\mathrm{EA}}$, the recorded observation images are arranged in temporal order as an image collage $I_t$ and provided to the VLM, allowing it to access the visual evidence acquired across multiple subgoal executions. If TAMP fails to find a feasible action sequence for a subgoal in $\mathcal{G}_t^{\mathrm{EA}}$, the system skips the remaining subgoals and proceeds to the next evidence-acquisition iteration with the observations collected so far and failure information.

During evidence acquisition, we represent the VLM's understanding of the environment as a textual evidence state:
\[
    E_t = (H_t, O_t, F_t),
\]
where $H_t$ is the executed-subgoal history, $O_t$ is a summary of observed facts, and $F_t$ is a set of retained findings relevant to later inference. The system records $H_t$, while the VLM generates and updates $O_t$ and $F_t$. The evidence state is maintained as external memory and provided to subsequent VLM calls.

The VLM updates the evidence state from $E_{t-1}$ to $E_t$ based on the observation collage $I_t$. The updated evidence state is then passed either to the next evidence-acquisition iteration or to the planning algorithm for generating task subgoals $\mathcal{G}^{\mathrm{task}}$. In this update, the VLM regenerates the textual components of the evidence state, namely $O_t$ and $F_t$. This allows outdated, redundant, or contradicted descriptions to be removed from the evidence state. EAFG incrementally refines its understanding of the environment during exploration while avoiding the generation of task subgoals $\mathcal{G}^{\mathrm{task}}$ that depend on unsupported assumptions.

\subsection{Feasibility Gating}
Feasibility Gating uses the evidence acquired in Section~\ref{subsec:ea} to determine whether to proceed with task planning, acquire further evidence, or halt. After executing the evidence-acquisition subgoals, we call the VLM to make this decision. Specifically, given the goal \(\mathcal{G}^{\mathrm{eng}}\), the preceding evidence state \(E_{t-1}\), and the observation collage \(I_t\), the VLM determines the feasibility status \(s_t\):
\[
    s_t \in \{\texttt{ready\_to\_plan}, \texttt{need\_more\_evidence}, \texttt{halt}\}.
\]
The VLM selects the status based on the decision criteria specified in the prompt. Each status determines the subsequent system behavior as follows.

For \(\texttt{ready\_to\_plan}\), the VLM judges that the available information supports a feasible complete task plan, without requiring further details or alternatives. When this status is selected, the system exits the evidence-acquisition loop and proceeds to evidence-conditioned task planning in Section~\ref{subsec:task_planning}.

For \(\texttt{need\_more\_evidence}\), the VLM determines that the available information is insufficient for a feasible task plan. Further inspection is justified by unexplored task-relevant areas or evidence suggesting where to obtain the missing information. In the same inference step, the VLM generates an additional sequence of evidence-acquisition subgoals \(\mathcal{G}_{t+1}^{\mathrm{EA}}\) to obtain the missing information. The system then returns to the evidence-acquisition loop in Section~\ref{subsec:ea}.

For \(\texttt{halt}\), the VLM judges that the task cannot proceed. Requirements in \(\mathcal{G}^{\mathrm{eng}}\) remain unverified after evidence acquisition, and the evidence provides no basis for further exploration. Upon selection, the system stops execution without proceeding to task-subgoal generation.

\subsection{Evidence-Conditioned Task Planning}
\label{subsec:task_planning}
Evidence-conditioned task planning uses the acquired evidence to generate task subgoals for the original goal instruction. The VLM receives the natural-language goal $\mathcal{G}^{\mathrm{eng}}$ and the accumulated evidence state $E_T$ as textual context, where $T$ denotes the evidence-acquisition iteration at which the gate selects \(\texttt{ready\_to\_plan}\). Given $\mathcal{G}^{\mathrm{eng}}$ and $E_T$, the VLM generates a sequence of task subgoals:
\[
    \mathcal{G}^{\mathrm{task}}
    = \left(g_1^{\mathrm{task}}, \ldots, g_M^{\mathrm{task}}\right),
\]
where $M$ is the number of generated task subgoals. Each task subgoal specifies a manipulation action aimed at fulfilling the instruction. The available actions include cooking and heating, which are excluded during evidence acquisition. For example, the VLM could generate subgoals to pick an observed ingredient, place it into a pot, and heat the pot.

Each generated task subgoal is passed to the TAMP module, which attempts to instantiate it as a geometrically and kinematically feasible action sequence. The robot executes the resulting sequence before proceeding to the next subgoal. If TAMP cannot find a feasible action sequence, the system discards the remaining task subgoals and returns to evidence acquisition in Section~\ref{subsec:ea}. In the next VLM call for evidence acquisition, the prompt includes the failed subgoal and available failure details, such as reported collisions and the robot's current holding state, to guide further exploration.

\begin{table*}[t]
\centering
\caption{\textbf{Experimental settings, research questions, and evaluation metrics.} Step and Recipe denote step and recipe completion rates. False Halt denotes the false halting rate in S1 and S2; Appropriate Halt denotes the appropriate halting rate in~S3.}
\label{tab:settings_rqs_metrics}
\setlength{\tabcolsep}{4pt}
\begin{tabular}{@{}>{\raggedright\arraybackslash}p{0.15\linewidth}
                  >{\raggedright\arraybackslash}p{0.23\linewidth}
                  >{\raggedright\arraybackslash}p{0.4\linewidth}
                  >{\raggedright\arraybackslash}p{0.15\linewidth}@{}}
\toprule
Setting & Goal & Research Question & Metrics \\
\midrule

S1: Explicit-and-present
&
\texttt{make chicken soup with salt and pepper}
&
\textbf{RQ1:} Can EAFG complete each recipe step and the full recipe when explicitly specified objects are present?
&
\parbox[t]{\linewidth}{\raggedright Step,\\Recipe,\\False Halt}
\\

\midrule

S2: Underspecified
&
\texttt{make chicken soup}
&
\textbf{RQ2:} Can EAFG acquire the required information through exploration and complete the same recipe under an underspecified goal?
&
\parbox[t]{\linewidth}{\raggedright Step,\\Recipe,\\False Halt}
\\

\midrule

S3: Explicit-but-absent
&
\texttt{make chicken soup with salt, pepper, and carrot}
&
\textbf{RQ3:} Can EAFG halt when a specified object is absent while reducing repeated subgoal attempts involving the absent object?
&
\parbox[t]{\linewidth}{\raggedright Appropriate Halt,\\Absent-object Subgoal Attempts}
\\

\bottomrule
\end{tabular}
\end{table*}

\section{Experiments}
\subsection{Kitchen Environment and Robot Embodiment}
We conduct experiments in a PyBullet kitchen simulation using a PR2 robot restricted to its left arm. Figure~\ref{fig:kitchen_environment} shows the initial kitchen state and object layout. The task requires long-horizon planning over object search beyond the initially visible area, access to storage spaces, placement of ingredients and seasonings, and the use of water and a stove. The environment consists of five movable objects, eight support surfaces such as counters, stove burners, a sink, and a pot, two storage spaces enclosed by doors or a drawer, and six articulated objects such as doors and knobs. Objects are added to the PDDL problem only after they appear in the robot's observation images. This environment was used in VLM-TAMP~\cite{vlm-tamp} as a benchmark for evaluating long-horizon task planning and geometric feasibility.

In our experiments, all doors are initially closed. The chicken leg is on the top refrigerator shelf, and the salt and pepper shakers are on the cabinet’s left and right sides, respectively, from the robot’s viewpoint. The robot is restricted to using only its left arm, so object rearrangement, storage-space access, and articulated-object manipulation must be performed within a limited reachable workspace. The robot can manipulate movable objects through pick-and-place actions and operate articulated joints by pulling, pushing, and rotating its wrist. Thus, successful execution requires jointly handling unobserved regions, closed storage spaces, and geometric constraints during cooking-related manipulation.

\subsection{Experimental Settings, Research Questions, and Evaluation Metrics}

We evaluate EAFG in three kitchen-task settings, each designed to correspond to a distinct research question, as summarized in Table~\ref{tab:settings_rqs_metrics}. These settings differ in whether the goal-relevant objects are explicitly specified in the instruction and whether those objects are actually present in the environment. This design allows us to evaluate three aspects of EAFG: completing a feasible recipe when the required objects are explicitly specified, acquiring missing information under an underspecified goal, and halting when an explicitly required object is absent.

S1 evaluates RQ1: whether the system can complete a feasible task when all explicitly specified goal-relevant objects are available. In this setting, the instruction specifies salt and pepper, and both objects are present in the environment. The step completion rate for each of the five recipe steps is the number of trials satisfying its completion criterion divided by the total number of trials. The five steps are placing the chicken leg in the pot, sprinkling salt and pepper into it (separately scored), adding water, and heating. Heating is counted once the pot contains both chicken and water on a stove with the corresponding knob on. The recipe completion rate is the number of trials completing all five steps in one trial divided by the total number of trials. We also report the false halting rate (False Halt), defined as the number of trials in which the VLM explicitly outputs \texttt{halt} divided by the total number of trials. Since this setting is designed to be feasible, such decisions are counted as false halts regardless of prior exploration. Termination due to exhaustion of the replanning or evidence-acquisition budget, or planning or execution failures is not counted as a false halt.

S2 evaluates RQ2: whether the system can acquire missing task-relevant information and complete the same recipe under an underspecified instruction. In this setting, the instruction asks the robot to make chicken soup but does not explicitly mention salt or pepper. We use the same step and recipe completion rates as in S1, including the salt and pepper steps. This evaluates whether the system can recover the information needed for the recipe through evidence acquisition rather than relying only on unsupported assumptions. We also report the false halting rate as in S1.

S3 evaluates RQ3: whether the system can recognize that a goal is infeasible when an explicitly required object is absent and avoid repeated attempts to process subgoals involving that absent object. In this setting, the instruction specifies carrot, but carrot is absent from the environment, so the goal cannot be achieved as specified. The appropriate halting rate is the number of trials in which the system halts after inspecting all predefined carrot source regions divided by the total number of trials. The four source regions are the fridge, the left and right compartments of the cabinet, and the pot body. These inspection requirements are used only for scoring and are not provided in the prompt. The absent-object subgoal attempts metric reports the mean number of times per trial that a VLM-generated subgoal referencing \texttt{carrot} is selected for processing. Although carrot is absent from the environment, the VLM may still generate subgoals referring to it. Each such processing attempt is counted, but the subgoal is rejected before TAMP planning because its object reference cannot be resolved.

\begin{table*}[t]
  \centering
  \caption{\textbf{S1: Explicit-and-present (RQ1) results} for the instruction \texttt{make chicken soup with salt and pepper}, comparing VLM-TAMP, VLM-TAMP-EH, and EAFG. Each condition uses 20 runs. Step and Recipe report step and recipe completion rates, respectively. False Halt reports the false halting rate; VLM-TAMP has no halt option. All rates lie in $[0,1]$.}
  \label{tab:chicken_salt_pepper_results}
  \setlength{\tabcolsep}{4pt}
  \begin{tabular}{llccccccc}
    \toprule
    VLM & Method & \multicolumn{5}{c}{Step $\uparrow$} & Recipe $\uparrow$ & False Halt $\downarrow$ \\
    \cmidrule(lr){3-7}
    & & Chicken & Water & Pepper & Salt & Stove & & \\
    \midrule
    \multirow{3}{*}{GPT-5.5} & VLM-TAMP~\cite{vlm-tamp} & \textbf{1.00} & 0.85 & \textbf{1.00} & \textbf{0.95} & 0.70 & 0.70 & \textbf{0.00} \\
     & VLM-TAMP-EH & 0.95 & 0.75 & 0.90 & \textbf{0.95} & 0.65 & 0.65 & \textbf{0.00} \\
     & EAFG (Ours) & 0.95 & \textbf{0.95} & 0.95 & \textbf{0.95} & \textbf{0.80} & \textbf{0.80} & 0.05 \\
    \midrule
    \multirow{3}{*}{Gemini-3.5-Flash} & VLM-TAMP~\cite{vlm-tamp} & 0.85 & \textbf{0.90} & 0.65 & 0.75 & 0.70 & 0.55 & \textbf{0.00} \\
     & VLM-TAMP-EH & \textbf{1.00} & \textbf{0.90} & 0.95 & \textbf{1.00} & 0.70 & 0.70 & 0.05 \\
     & EAFG (Ours) & \textbf{1.00} & \textbf{0.90} & \textbf{1.00} & \textbf{1.00} & \textbf{0.85} & \textbf{0.85} & \textbf{0.00} \\
    \bottomrule
  \end{tabular}
\end{table*}

\begin{table*}[t]
  \centering
  \caption{\textbf{S2: Underspecified (RQ2) results} for the instruction \texttt{make chicken soup}, comparing VLM-TAMP, VLM-TAMP-EH, and EAFG. Each condition uses 20 runs. Step and Recipe report step and recipe completion rates, respectively. False Halt reports the false halting rate; VLM-TAMP has no halt option. All rates lie in $[0,1]$.}
  \label{tab:cc_model_chicken_soup_results}
  \setlength{\tabcolsep}{4pt}
  \begin{tabular}{llccccccc}
    \toprule
    VLM & Method & \multicolumn{5}{c}{Step $\uparrow$} & Recipe $\uparrow$ & False Halt $\downarrow$ \\
    \cmidrule(lr){3-7}
    & & Chicken & Water & Pepper & Salt & Stove & & \\
    \midrule
    \multirow{3}{*}{GPT-5.5} & VLM-TAMP~\cite{vlm-tamp} & \textbf{1.00} & 0.95 & 0.30 & 0.30 & 0.85 & 0.15 & \textbf{0.00} \\
     & VLM-TAMP-EH & 0.80 & 0.75 & 0.65 & 0.70 & 0.65 & 0.55 & \textbf{0.00} \\
     & EAFG (Ours) & \textbf{1.00} & \textbf{1.00} & \textbf{1.00} & \textbf{1.00} & \textbf{0.95} & \textbf{0.95} & \textbf{0.00} \\
    \midrule
    \multirow{3}{*}{Gemini-3.5-Flash} & VLM-TAMP~\cite{vlm-tamp} & \textbf{1.00} & \textbf{1.00} & 0.00 & 0.10 & 0.95 & 0.00 & \textbf{0.00} \\
     & VLM-TAMP-EH & \textbf{1.00} & \textbf{1.00} & 0.05 & 0.25 & \textbf{1.00} & 0.05 & 0.15 \\
     & EAFG (Ours) & 0.90 & 0.90 & \textbf{0.40} & \textbf{0.75} & 0.85 & \textbf{0.35} & \textbf{0.00} \\
    \bottomrule
  \end{tabular}
\end{table*}

\subsection{Baselines}
We compare EAFG with the following two baselines.

\paragraph{VLM-TAMP\cite{vlm-tamp}}
It generates a sequence of task subgoals from a natural-language instruction and visual observations. Each subgoal is passed to TAMP to obtain an executable action sequence. When planning fails, the VLM replans the remaining subgoals based on the available observations and execution history. Its prompt does not include a halt option, so it cannot explicitly halt when the task is infeasible. We therefore do not evaluate VLM-TAMP in S3 and mark its results with ``-'' in Table~\ref{tab:carrot_eafg_model_results}.

\paragraph{VLM-TAMP-EH (Exploration and Halting)}
It augments VLM-TAMP with explicit instructions in the prompt to explore for missing task-relevant information and halt when observations and history indicate infeasibility. Unlike EAFG, it incorporates exploration into ordinary subgoal generation without a separate evidence-acquisition phase or an explicit feasibility gate before task planning.

\subsection{Implementation Details}
We use \texttt{gpt-5.5-\allowbreak 2026-\allowbreak 04-\allowbreak 23} and \texttt{gemini-3.5-\allowbreak flash} as VLMs, evaluating them independently. Within each model condition, evidence acquisition and gating, task planning, and predicate translation use the same VLM with different prompts. The VLM returns a JSON object containing the gate status, updated observations and findings, and any additional exploration subgoals during evidence acquisition. During task planning, it returns a sequence of subgoals in English. Each generated subgoal sequence is translated into predicate strings by an additional VLM call using a predefined vocabulary of object names and types. We sample one response per call, using a temperature of $0.2$ for evidence acquisition, gating, and task planning and $0$ for predicate translation. In EAFG, we allow up to five evidence acquisition iterations. For both the baselines and EAFG, we allow up to five replanning attempts by the VLM in response to task planning failures.

\begin{figure*}[t]
  \centering
  \includegraphics[width=0.95\linewidth]{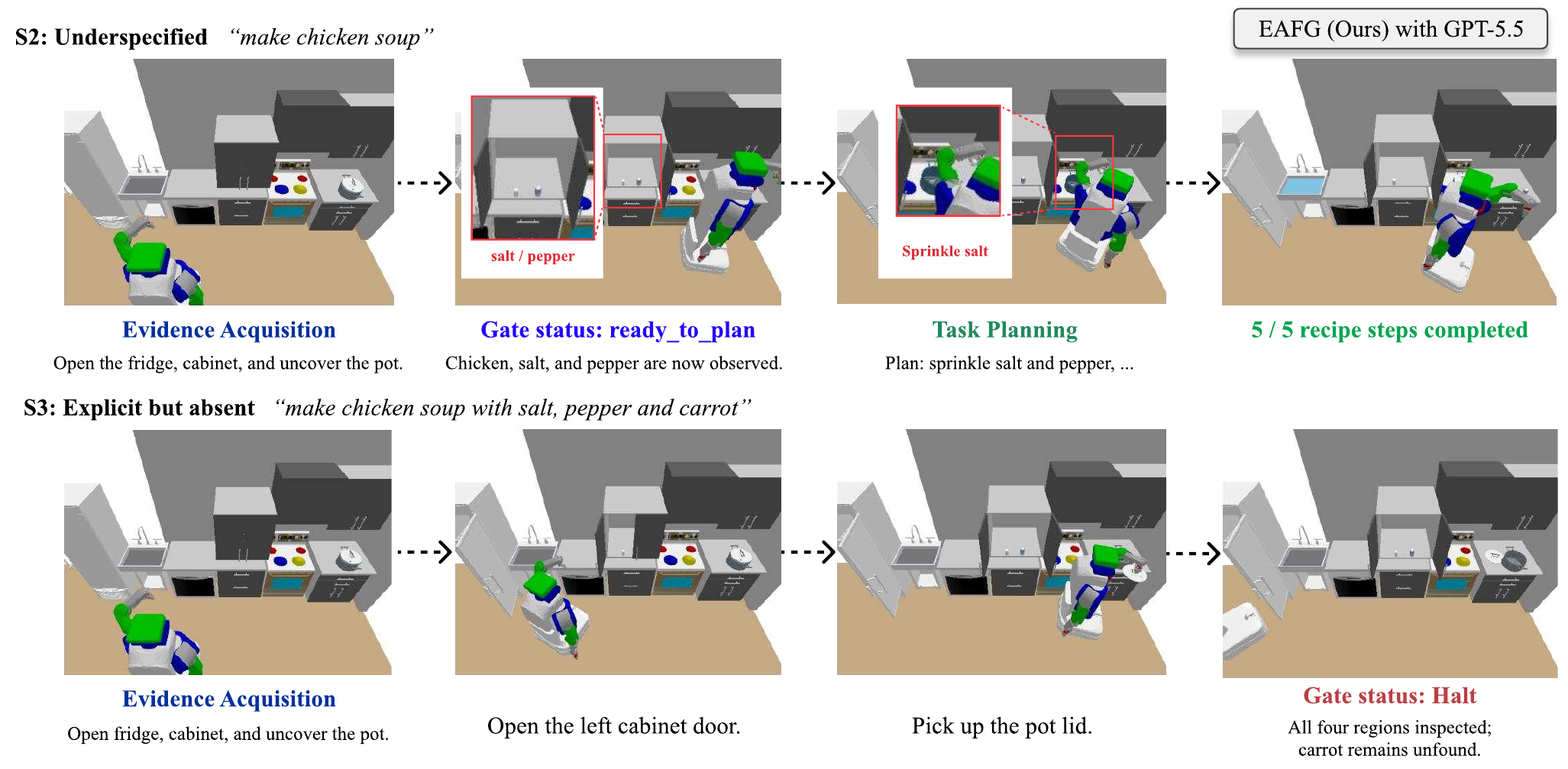}
  \caption{Example EAFG trajectories with GPT-5.5 in S2 (top) and S3 (bottom). Evidence acquisition supports recipe completion under an underspecified instruction and halting when carrot is absent. Selected keyframes are shown; intermediate exploration and replanning are omitted.}
  \label{fig:qualitative_results}
\end{figure*}

Predicate strings are converted into PDDL goals by resolving object names to their corresponding identifiers. TAMP supports base motion, picking, placement, sprinkling, and joint manipulation, with manipulation operators parameterized by their target objects. For TAMP, we use PDDLStream~\cite{pddlstream} in diverse planning mode and consider up to 12 plan skeletons. If planning fails for a subgoal, we replan up to three times while gradually increasing the set of observed objects included in planning. Planning is considered successful if any attempt yields an executable action sequence.

\subsection{Results}
We compare EAFG with VLM-TAMP and VLM-TAMP-EH in S1 and S2, and with VLM-TAMP-EH in S3. EAFG achieves the highest recipe completion rate for both VLMs in S1 and S2. The largest gains occur in S2, where the instruction does not explicitly mention salt and pepper. In S3, EAFG increases the appropriate halting rate and reduces subgoal attempts involving the absent carrot.

\paragraph{S1: Explicit-and-present.}
Table~\ref{tab:chicken_salt_pepper_results} shows that EAFG achieves the highest recipe completion rate for both VLMs, reaching 0.80 for GPT-5.5 and 0.85 for Gemini-3.5-Flash. The gains over both baselines suggest that acquiring evidence before task planning helps complete recipes when the required ingredients are explicitly specified. False halting also remains possible even when all required objects are present, although it occurs infrequently in S1: GPT-5.5 with EAFG and Gemini-3.5-Flash with VLM-TAMP-EH each produce one false halt (0.05).

Gemini-3.5-Flash slightly exceeds GPT-5.5 in recipe completion with both VLM-TAMP-EH (0.70 versus 0.65) and EAFG (0.85 versus 0.80), although each difference corresponds to only one trial out of 20. With VLM-TAMP, this ordering reverses: Gemini-3.5-Flash achieves 0.55, below GPT-5.5's 0.70 and the lowest recipe completion rate among the S1 conditions. The logs show that some replanned subgoal sequences omit one or both seasoning steps. Without explicit exploration guidance, replanning may favor already observed objects and omit searches for unobserved seasonings, contributing to the lower completion rate.

\paragraph{S2: Underspecified.}

Table~\ref{tab:cc_model_chicken_soup_results} shows that EAFG achieves the highest recipe completion rate for both VLMs in S2, followed by VLM-TAMP-EH and VLM-TAMP. Its gains over both baselines are larger than in S1, suggesting that evidence acquisition is particularly beneficial for recipe completion under underspecified instructions. False halts occur only for Gemini-3.5-Flash with VLM-TAMP-EH, in three trials (0.15), whereas EAFG has no false halts with either VLM.

In S2, the instruction does not explicitly mention pepper or salt. Compared with S1, both seasoning completion rates decline substantially for both baselines and for EAFG with Gemini-3.5-Flash, with failures in either seasoning step preventing completion of the full recipe.

For GPT-5.5, recipe completion increases from 0.15 with VLM-TAMP to 0.55 with VLM-TAMP-EH and 0.95 with EAFG, an absolute gain of 0.40 at each transition. The first gain suggests that explicitly prompting exploration can help include seasoning steps under an underspecified instruction. The further gain with EAFG accompanies completion rates of 1.00 for both pepper and salt, suggesting that the acquired evidence helps the VLM formulate concrete seasoning subgoals. Only GPT-5.5 with EAFG achieves higher recipe completion in S2, increasing from 0.80 in S1 to 0.95. Since heating is incomplete more often than either seasoning step in S1, differences in manipulation planning and recovery through replanning after failures may also contribute to this variation between settings. In the S2 example in Fig.~\ref{fig:qualitative_results} (top), EAFG observes salt and pepper, selects \texttt{ready\_to\_plan}, and completes all five recipe steps.

Gemini-3.5-Flash has lower recipe completion rates than GPT-5.5 across all three methods: 0.00 with VLM-TAMP, 0.05 with VLM-TAMP-EH, and 0.35 with EAFG, with most incomplete recipes lacking at least one seasoning step. EAFG improves both seasoning completion rates over the baselines, but pepper completion remains lower than salt completion (0.40 versus 0.75). In 17 of the 20 EAFG trials, Gemini-3.5-Flash opens the left cabinet door first. After finding salt on the left, it sometimes treats the cabinet inspection as complete and omits further exploration of the right side, where pepper is located, potentially contributing to the lower pepper completion rate. The two EAFG trials without chicken placement ended when the evidence-acquisition budget was exhausted.

\paragraph{S3: Explicit-but-absent.}
\begin{table*}[t]
  \centering
  \caption{\textbf{S3: Explicit-but-absent (RQ3) results} for \texttt{make chicken soup with salt, pepper, and carrot}, with carrot absent. We compare VLM-TAMP-EH and EAFG over 20 runs per condition; VLM-TAMP is unevaluated (``-'') because its prompt lacks a halt option. Appropriate Halt is the rate ($[0,1]$) of halting after inspecting all predefined source regions, including the pot body. Absent-object Subgoal Attempts reports mean per-trial attempts to process subgoals involving the absent object.}
  \label{tab:carrot_eafg_model_results}
  \begin{tabular}{llcc}
    \toprule
    VLM & Method & Appropriate Halt $\uparrow$ & \begin{tabular}[c]{@{}c@{}}Absent-object\\Subgoal Attempts $\downarrow$\end{tabular} \\
    \midrule
    \multirow{3}{*}{GPT-5.5} & VLM-TAMP~\cite{vlm-tamp} & - & - \\
     & VLM-TAMP-EH & 0.15 & 4.95 \\
     & EAFG (Ours) & \textbf{0.95} & \textbf{0.00} \\
    \midrule
    \multirow{3}{*}{Gemini-3.5-Flash} & VLM-TAMP~\cite{vlm-tamp} & - & - \\
     & VLM-TAMP-EH & 0.25 & 3.55 \\
     & EAFG (Ours) & \textbf{0.80} & \textbf{0.35} \\
    \bottomrule
  \end{tabular}
\end{table*}
Table~\ref{tab:carrot_eafg_model_results} compares EAFG with VLM-TAMP-EH in S3, where the instruction explicitly requires carrot but no carrot is present in the environment. For GPT-5.5, EAFG increases the appropriate halting rate from 0.15 to 0.95 and reduces the mean number of absent-object subgoal attempts per trial from 4.95 to 0.00. For Gemini-3.5-Flash, EAFG increases the appropriate halting rate from 0.25 to 0.80 and reduces the mean number of absent-object subgoal attempts per trial from 3.55 to 0.35. Across both VLMs, EAFG more often halts after inspecting all predefined carrot source regions and less frequently selects subgoals referencing the absent carrot for processing. In the S3 example in Fig.~\ref{fig:qualitative_results} (bottom), EAFG inspects all four source regions and selects \texttt{halt}, without attempting a carrot-related subgoal.

\section{Limitations}
Our controlled simulation study evaluates the planning-level effectiveness of EAFG by controlling the visibility and availability of goal-relevant objects. Generalization to real-world images and end-to-end robustness under real-world perception and manipulation errors remain unvalidated.

EAFG uses TAMP planning failures to guide further exploration. Future work includes integrating execution feedback to verify whether planned actions succeed on physical robots and to support recovery from manipulation failures.

\section{Conclusion}
We proposed EAFG, a framework that acquires observational evidence to specify task plans consistent with the environment and uses the acquired evidence to determine whether to proceed with task planning, acquire further evidence, or halt. Experiments across three cooking scenarios show that EAFG improves task completion over VLM-TAMP baselines for both explicitly specified and underspecified instructions. EAFG also halts appropriately when a required object is absent from the environment. These results support the use of observational evidence both to identify appropriate task subgoals and to assess task feasibility.

\bibliographystyle{IEEEtran}
\bibliography{example}

\begin{thebibliography}{10}
\providecommand{\url}[1]{#1}
\csname url@rmstyle\endcsname
\providecommand{\newblock}{\relax}
\providecommand{\bibinfo}[2]{#2}
\providecommand\BIBentrySTDinterwordspacing{\spaceskip=0pt\relax}
\providecommand\BIBentryALTinterwordstretchfactor{4}
\providecommand\BIBentryALTinterwordspacing{\spaceskip=\fontdimen2\font plus
\BIBentryALTinterwordstretchfactor\fontdimen3\font minus
  \fontdimen4\font\relax}
\providecommand\BIBforeignlanguage[2]{{%
\expandafter\ifx\csname l@#1\endcsname\relax
\typeout{** WARNING: IEEEtran.bst: No hyphenation pattern has been}%
\typeout{** loaded for the language `#1'. Using the pattern for}%
\typeout{** the default language instead.}%
\else
\language=\csname l@#1\endcsname
\fi
#2}}

\bibitem{tamp}
S.~Srivastava, E.~Fang, L.~Riano, R.~Chitnis, S.~Russell, and P.~Abbeel,
  ``Combined task and motion planning through an extensible planner-independent
  interface layer,'' in \emph{2014 IEEE International Conference on Robotics
  and Automation (ICRA)}, 2014, pp. 639--646.

\bibitem{pddlstream}
C.~R. Garrett, T.~Lozano-Perez, and L.~P. Kaelbling, ``{PDDLStream}:
  Integrating symbolic planners and blackbox samplers via optimistic adaptive
  planning,'' in \emph{Proceedings of the International Conference on Automated
  Planning and Scheduling}, vol.~30, 2020, pp. 440--448.

\bibitem{vlm-tamp}
Z.~Yang, C.~Garrett, D.~Fox, T.~Lozano-P{\'e}rez, and L.~P. Kaelbling,
  ``Guiding long-horizon task and motion planning with vision language
  models,'' in \emph{2025 IEEE International Conference on Robotics and
  Automation (ICRA)}.\hskip 1em plus 0.5em minus 0.4em\relax IEEE, 2025, pp.
  16\,847--16\,853.

\bibitem{owl-tamp}
N.~Kumar, W.~Shen, F.~Ramos, D.~Fox, T.~Lozano-Pérez, L.~P. Kaelbling, and
  C.~R. Garrett, ``Open-world task and motion planning via vision-language
  model generated constraints,'' \emph{IEEE Robotics and Automation Letters},
  vol.~11, no.~3, pp. 3366--3373, 2026.

\bibitem{lee-etal-2025-vlind}
K.-i. Lee, M.~Kim, S.~Yoon, M.~Kim, D.~Lee, H.~Koh, and K.~Jung,
  ``{VLind-Bench}: Measuring language priors in large vision-language models,''
  in \emph{Findings of NAACL}, 2025, pp. 4129--4144.

\bibitem{guan2024hallusionbench}
T.~Guan, F.~Liu, X.~Wu, R.~Xian, Z.~Li, X.~Liu, X.~Wang, L.~Chen, F.~Huang,
  Y.~Yacoob, \emph{et~al.}, ``{HallusionBench}: an advanced diagnostic suite
  for entangled language hallucination and visual illusion in large
  vision-language models,'' in \emph{Computer Vision and Pattern Recognition
  (CVPR)}, 2024, pp. 14\,375--14\,385.

\bibitem{garrett2021integrated}
C.~R. Garrett, R.~Chitnis, R.~Holladay, B.~Kim, T.~Silver, L.~P. Kaelbling, and
  T.~Lozano-P{\'e}rez, ``Integrated task and motion planning,'' \emph{Annual
  Review of Control, Robotics, and Autonomous Systems}, vol.~4, no.~1, pp.
  265--293, 2021.

\bibitem{kaelbling2011hierarchical}
L.~P. Kaelbling and T.~Lozano-P{\'e}rez, ``Hierarchical task and motion
  planning in the now,'' in \emph{2011 IEEE International Conference on
  Robotics and Automation}, 2011, pp. 1470--1477.

\bibitem{marc2015logic}
M.~Toussaint, ``Logic-geometric programming: an optimization-based approach to
  combined task and motion planning,'' in \emph{Proceedings of the 24th
  International Conference on Artificial Intelligence}, 2015, pp. 1930--1936.

\bibitem{pmlr-v205-ichter23a}
B.~Ichter, A.~Brohan, Y.~Chebotar, C.~Finn, K.~Hausman, A.~Herzog, D.~Ho,
  J.~Ibarz, A.~Irpan, E.~Jang, R.~Julian, D.~Kalashnikov, S.~Levine, Y.~Lu,
  C.~Parada, K.~Rao, P.~Sermanet, A.~T. Toshev, V.~Vanhoucke, F.~Xia, T.~Xiao,
  P.~Xu, M.~Yan, N.~Brown, M.~Ahn, O.~Cortes, N.~Sievers, C.~Tan, S.~Xu,
  D.~Reyes, J.~Rettinghouse, J.~Quiambao, P.~Pastor, L.~Luu, K.-H. Lee,
  Y.~Kuang, S.~Jesmonth, N.~J. Joshi, K.~Jeffrey, R.~J. Ruano, J.~Hsu,
  K.~Gopalakrishnan, B.~David, A.~Zeng, and C.~K. Fu, ``Do as {I} can, not as
  {I} say: Grounding language in robotic affordances,'' in \emph{Proceedings of
  The 6th Conference on Robot Learning}, 2023, pp. 287--318.

\bibitem{liang2023code}
J.~Liang, W.~Huang, F.~Xia, P.~Xu, K.~Hausman, B.~Ichter, P.~Florence, and
  A.~Zeng, ``Code as policies: Language model programs for embodied control,''
  in \emph{2023 IEEE International Conference on Robotics and Automation
  (ICRA)}, 2023, pp. 9493--9500.

\bibitem{10161317}
I.~Singh, V.~Blukis, A.~Mousavian, A.~Goyal, D.~Xu, J.~Tremblay, D.~Fox,
  J.~Thomason, and A.~Garg, ``{ProgPrompt}: Generating situated robot task
  plans using large language models,'' in \emph{2023 IEEE International
  Conference on Robotics and Automation (ICRA)}, 2023, pp. 11\,523--11\,530.

\bibitem{huang2023voxposer}
\BIBentryALTinterwordspacing
W.~Huang, C.~Wang, R.~Zhang, Y.~Li, J.~Wu, and L.~Fei-Fei, ``{VoxPoser}:
  Composable {3D} value maps for robotic manipulation with language models,''
  in \emph{7th Annual Conference on Robot Learning}, 2023. [Online]. Available:
  \url{https://openreview.net/forum?id=9_8LF30mOC}
\BIBentrySTDinterwordspacing

\bibitem{liu2023llmp}
B.~Liu, Y.~Jiang, X.~Zhang, Q.~Liu, S.~Zhang, J.~Biswas, and P.~Stone,
  ``{LLM+P}: Empowering large language models with optimal planning
  proficiency,'' \emph{arXiv preprint arXiv:2304.11477}, 2023.

\bibitem{chen2024autotamp}
Y.~Chen, J.~Arkin, C.~Dawson, Y.~Zhang, N.~Roy, and C.~Fan, ``{AutoTAMP}:
  Autoregressive task and motion planning with {LLM}s as translators and
  checkers,'' in \emph{2024 IEEE International Conference on Robotics and
  Automation (ICRA)}, 2024, pp. 6695--6702.

\bibitem{guo2025castl}
W.~Guo, Z.~Kingston, and L.~E. Kavraki, ``{CaStL}: Constraints as
  specifications through {LLM} translation for long-horizon task and motion
  planning,'' in \emph{2025 IEEE International Conference on Robotics and
  Automation (ICRA)}, 2025, pp. 11\,957--11\,964.

\bibitem{athalye2025pixelspredicateslearningsymbolic}
A.~Athalye, N.~Kumar, T.~Silver, Y.~Liang, J.~Wang, T.~Lozano-Pérez, and L.~P.
  Kaelbling, ``From pixels to predicates: Learning symbolic world models via
  pretrained {VLM}s,'' \emph{IEEE Robotics and Automation Letters (RA-L)},
  vol.~11, pp. 4002--4009, 2026.

\bibitem{li2016act}
J.~K. Li, D.~Hsu, and W.~S. Lee, ``Act to see and see to act: {POMDP} planning
  for objects search in clutter,'' in \emph{2016 IEEE/RSJ International
  Conference on Intelligent Robots and Systems (IROS)}, 2016, pp. 5701--5707.

\bibitem{xiao2019online}
Y.~Xiao, S.~Katt, A.~ten Pas, S.~Chen, and C.~Amato, ``Online planning for
  target object search in clutter under partial observability,'' in \emph{2019
  International Conference on Robotics and Automation (ICRA)}, 2019, pp.
  8241--8247.

\bibitem{kaelbling2013integrated}
L.~P. Kaelbling and T.~Lozano-P{\'e}rez, ``Integrated task and motion planning
  in belief space,'' \emph{The International Journal of Robotics Research},
  vol.~32, no. 9-10, pp. 1194--1227, 2013.

\bibitem{ss-replan}
C.~R. Garrett, C.~Paxton, T.~Lozano-Pérez, L.~P. Kaelbling, and D.~Fox,
  ``Online replanning in belief space for partially observable task and motion
  problems,'' in \emph{2020 IEEE International Conference on Robotics and
  Automation (ICRA)}, 2020, pp. 5678--5684.

\bibitem{curtis2024partially}
A.~Curtis, G.~Matheos, N.~Gothoskar, V.~Mansinghka, J.~Tenenbaum,
  T.~Lozano-Pérez, and L.~P. Kaelbling, ``Partially observable task and motion
  planning with uncertainty and risk awareness,'' 2024.

\bibitem{paudel2026objectsearch}
A.~Paudel, A.~Khanal, R.~I. Arnob, S.~Hossain, and G.~J. Stein, ``Object search
  in partially-known environments via {LLM}-informed model-based planning and
  prompt selection,'' 2026.

\bibitem{honerkamp2024language}
D.~Honerkamp, M.~B{\"u}chner, F.~Despinoy, T.~Welschehold, and A.~Valada,
  ``Language-grounded dynamic scene graphs for interactive object search with
  mobile manipulation,'' \emph{IEEE Robotics and Automation Letters}, vol.~9,
  no.~10, pp. 8298--8305, 2024.

\end{thebibliography}

\end{document}